\documentclass[conference,10pt]{./MetaFiles/IEEEtran}

\IEEEoverridecommandlockouts

\usepackage{gensymb}
\usepackage{dsfont}
\usepackage{amsmath,amssymb,amsfonts,amsthm}

\usepackage{epsfig}
\usepackage{cite}
\usepackage{hhline}
\usepackage{multirow}
\usepackage{xcolor}
\usepackage{makecell}
\usepackage{subcaption}
\usepackage{capt-of}
\usepackage[labelformat=simple]{subcaption}

\usepackage{enumerate}
\usepackage{enumitem}
\usepackage{color}
\usepackage{graphicx}
\usepackage{algpseudocode}
\usepackage[ruled]{algorithm}
\makeatletter
\newcounter{parentalgorithm}

\makeatother

\usepackage[nolist,printonlyused]{acronym}      
\usepackage{booktabs}
\usepackage{bm}
\usepackage{pgf}
\usepackage{pgfplots}
\usepackage{tikz}
\usepackage{grffile}
\pgfplotsset{compat=newest}
\usetikzlibrary{plotmarks}
\usetikzlibrary{shapes.geometric,quotes,angles,automata,arrows,positioning,calc,3d,matrix,decorations.markings,matrix}

\tikzstyle{startstop} = [rectangle, rounded corners, minimum width=3cm, minimum height=1cm,text centered, draw=black, fill=white!30]
\tikzstyle{io} = [trapezium, trapezium left angle=70, trapezium right angle=110, minimum width=3cm, minimum height=1cm, text centered, draw=black, fill=white!30]
\tikzstyle{process} = [rectangle, minimum width=2cm, minimum height=1cm, text centered, draw=black, fill=white!30]
\tikzstyle{decision} = [diamond, minimum width=3cm, minimum height=1cm, text centered, draw=black, fill=white!30]
\tikzstyle{arrow} = [thick,->,>=stealth]

\makeatletter
\newcommand{\vast}{\bBigg@{3}}
\newcommand{\Vast}{\bBigg@{4}}
\makeatother
\renewcommand{\IEEEQED}{\IEEEQEDopen}

\showoutput
\makeatletter
\tikzset{%
  remember picture with id/.style={%
    remember picture,
    overlay,
    save picture id=#1,
  },
  save picture id/.code={%
    \edef\pgf@temp{#1}%
    \immediate\write\pgfutil@auxout{%
      \noexpand\savepointas{\pgf@temp}{\pgfpictureid}}%
  },
  if picture id/.code args={#1#2#3}{%
    \@ifundefined{save@pt@#1}{%
      \pgfkeysalso{#3}%
    }{
      \pgfkeysalso{#2}%
    }
  }
}

\def\savepointas#1#2{%
  \expandafter\gdef\csname save@pt@#1\endcsname{#2}%
}

\def\tmk@labeldef#1,#2\@nil{%
  \def\tmk@label{#1}%
  \def\tmk@def{#2}%
}

\tikzdeclarecoordinatesystem{pic}{%
  \pgfutil@in@,{#1}%
  \ifpgfutil@in@%
    \tmk@labeldef#1\@nil
  \else
    \tmk@labeldef#1,(0pt,0pt)\@nil
  \fi
  \@ifundefined{save@pt@\tmk@label}{%
    \tikz@scan@one@point\pgfutil@firstofone\tmk@def
  }{%
  \pgfsys@getposition{\csname save@pt@\tmk@label\endcsname}\save@orig@pic%
  \pgfsys@getposition{\pgfpictureid}\save@this@pic%
  \pgf@process{\pgfpointorigin\save@this@pic}%
  \pgf@xa=\pgf@x
  \pgf@ya=\pgf@y
  \pgf@process{\pgfpointorigin\save@orig@pic}%
  \advance\pgf@x by -\pgf@xa
  \advance\pgf@y by -\pgf@ya
  }%
}

\makeatother

\usepackage{comment}
\usepackage{xpatch}
\makeatletter
\xpatchcmd{\algorithmic}{\itemsep\z@}{\itemsep=-0.25mm}{}{}
\makeatother

\usepackage{matlab-prettifier}

\usepackage{multicol}
\usepackage{multirow}

\usepackage{xcolor} 

\usepackage{amsmath}
\usepackage{algorithm}

\algrenewcommand\algorithmicforall{\textbf{foreach}}
\algrenewcommand\algorithmicindent{.8em}

\usepackage{pgfplots}
\usetikzlibrary{pgfplots.groupplots}

\usepackage{mathtools}

\DeclarePairedDelimiter\floor{\lfloor}{\rfloor}
\begin{document}

\title{Joint Energy and Latency Optimization in Federated Learning over Cell-Free Massive MIMO Networks}

\author{Afsaneh Mahmoudi, Mahmoud Zaher, and Emil Björnson \\
 School of Electrical Engineering and Computer Science \\KTH Royal Institute of Technology, Stockholm, Sweden \\
Emails: \{afmb, mahmoudz, emilbjo\}@kth.se 
\thanks{The work was supported by the SUCCESS grant from the Swedish Foundation for Strategic Research.}
} 


\newtheorem{theorem}{Theorem}
\newtheorem{defin}{Definition}
\newtheorem{prop}{Proposition}
\newtheorem{lemma}{Lemma}
\newtheorem{corollary}{Corollary}
\newtheorem{alg}{Algorithm}
\newtheorem{remark}{Remark}
\newtheorem{result}{Result}
\newtheorem{conjecture}{Conjecture}
\newtheorem{example}{Example}
\newtheorem{notations}{Notations}
\newtheorem{assumption}{Assumption}
\newcommand{\combin}[2]{\ensuremath{ \left( \ba{c} #1 \\ #2 \ea \right) }}
\newcommand{\diag}{{\mbox{diag}}}
\newcommand{\rank}{{\mbox{rank}}}
\newcommand{\dom}{{\mbox{dom{\color{white!100!black}.}}}}
\newcommand{\range}{{\mbox{range{\color{white!100!black}.}}}}
\newcommand{\image}{{\mbox{image{\color{white!100!black}.}}}}
\newcommand{\herm}{^{\mbox{\scriptsize H}}}  
\newcommand{\sherm}{^{\mbox{\tiny H}}}       
\newcommand{\tran}{^{\mbox{\scriptsize T}}}  
\newcommand{\tranIn}{^{\mbox{-\scriptsize T}}}  
\newcommand{\card}{{\mbox{\textbf{card}}}}
\newcommand{\asign}{{\mbox{$\colon\hspace{-2mm}=\hspace{1mm}$}}}
\newcommand{\ssum}[1]{\mathop{ \textstyle{\sum}}_{#1}}

\newcommand{\vbar}{\raisebox{.17ex}{\rule{.04em}{1.35ex}}}
\newcommand{\vbarind}{\raisebox{.01ex}{\rule{.04em}{1.1ex}}}
\newcommand{\D}{\ifmmode {\rm I}\hspace{-.2em}{\rm D} \else ${\rm I}\hspace{-.2em}{\rm D}$ \fi}
\newcommand{\T}{\ifmmode {\rm I}\hspace{-.2em}{\rm T} \else ${\rm I}\hspace{-.2em}{\rm T}$ \fi}
\newcommand{\B}{\ifmmode {\rm I}\hspace{-.2em}{\rm B} \else \mbox{${\rm I}\hspace{-.2em}{\rm B}$} \fi}
\newcommand{\Hil}{\ifmmode {\rm I}\hspace{-.2em}{\rm H} \else \mbox{${\rm I}\hspace{-.2em}{\rm H}$} \fi}
\newcommand{\C}{\ifmmode \hspace{.2em}\vbar\hspace{-.31em}{\rm C} \else \mbox{$\hspace{.2em}\vbar\hspace{-.31em}{\rm C}$} \fi}
\newcommand{\Cind}{\ifmmode \hspace{.2em}\vbarind\hspace{-.25em}{\rm C} \else \mbox{$\hspace{.2em}\vbarind\hspace{-.25em}{\rm C}$} \fi}
\newcommand{\Q}{\ifmmode \hspace{.2em}\vbar\hspace{-.31em}{\rm Q} \else \mbox{$\hspace{.2em}\vbar\hspace{-.31em}{\rm Q}$} \fi}
\newcommand{\Z}{\ifmmode {\rm Z}\hspace{-.28em}{\rm Z} \else ${\rm Z}\hspace{-.38em}{\rm Z}$ \fi}

\newcommand{\sgn}{\mbox {sgn}}
\newcommand{\var}{\mbox {var}}
\newcommand{\E}{\mbox {E}}
\newcommand{\cov}{\mbox {cov}}
\renewcommand{\Re}{\mbox {Re}}
\renewcommand{\Im}{\mbox {Im}}
\newcommand{\cum}{\mbox {cum}}

\renewcommand{\vec}[1]{{\bf{#1}}}     

\newcommand{\vecsc}[1]{\mbox {\boldmath \scriptsize $#1$}}     
\newcommand{\itvec}[1]{\mbox {\boldmath $#1$}}
\newcommand{\itvecsc}[1]{\mbox {\boldmath $\scriptstyle #1$}}
\newcommand{\gvec}[1]{\mbox{\boldmath $#1$}}

\newcommand{\balpha}{\mbox {\boldmath $\alpha$}}
\newcommand{\bbeta}{\mbox {\boldmath $\beta$}}
\newcommand{\bgamma}{\mbox {\boldmath $\gamma$}}
\newcommand{\bdelta}{\mbox {\boldmath $\delta$}}
\newcommand{\bepsilon}{\mbox {\boldmath $\epsilon$}}
\newcommand{\bvarepsilon}{\mbox {\boldmath $\varepsilon$}}
\newcommand{\bzeta}{\mbox {\boldmath $\zeta$}}
\newcommand{\boldeta}{\mbox {\boldmath $\eta$}}
\newcommand{\btheta}{\mbox {\boldmath $\theta$}}
\newcommand{\bvartheta}{\mbox {\boldmath $\vartheta$}}
\newcommand{\biota}{\mbox {\boldmath $\iota$}}
\newcommand{\blambda}{\mbox {\boldmath $\lambda$}}
\newcommand{\bmu}{\mbox {\boldmath $\mu$}}
\newcommand{\bnu}{\mbox {\boldmath $\nu$}}
\newcommand{\bxi}{\mbox {\boldmath $\xi$}}
\newcommand{\bpi}{\mbox {\boldmath $\pi$}}
\newcommand{\bvarpi}{\mbox {\boldmath $\varpi$}}
\newcommand{\brho}{\mbox {\boldmath $\rho$}}
\newcommand{\bvarrho}{\mbox {\boldmath $\varrho$}}
\newcommand{\bsigma}{\mbox {\boldmath $\sigma$}}
\newcommand{\bvarsigma}{\mbox {\boldmath $\varsigma$}}
\newcommand{\btau}{\mbox {\boldmath $\tau$}}
\newcommand{\bupsilon}{\mbox {\boldmath $\upsilon$}}
\newcommand{\bphi}{\mbox {\boldmath $\phi$}}
\newcommand{\bvarphi}{\mbox {\boldmath $\varphi$}}
\newcommand{\bchi}{\mbox {\boldmath $\chi$}}
\newcommand{\bpsi}{\mbox {\boldmath $\psi$}}
\newcommand{\bomega}{\mbox {\boldmath $\omega$}}

\newcommand{\R}{\mathbb{R}}
\newcommand{\N}{\mathbb{N}}

\def\calA{{\mathcal A}}
\def\calB{{\mathcal B}}
\def\calC{{\mathcal C}}
\def\calD{{\mathcal D}}
\def\calE{{\mathcal E}}
\def\calF{{\mathcal F}}
\def\calG{{\mathcal G}}
\def\calH{{\mathcal H}}
\def\calI{{\mathcal I}}
\def\calJ{{\mathcal J}}
\def\calK{{\mathcal K}}
\def\calL{{\mathcal L}}
\def\calM{{\mathcal M}}
\def\calN{{\mathcal N}}
\def\calO{{\mathcal O}}
\def\calP{{\mathcal P}}
\def\calQ{{\mathcal Q}}
\def\calR{{\mathcal R}}
\def\calS{{\mathcal S}}
\def\calT{{\mathcal T}}
\def\calU{{\mathcal U}}
\def\calV{{\mathcal V}}
\def\calW{{\mathcal W}}
\def\calX{{\mathcal X}}
\def\calY{{\mathcal Y}}
\def\calZ{{\mathcal Z}}

\def\bA{\mbox {\boldmath $A$}}
\def\bB{\mbox {\boldmath $B$}}
\def\bC{\mbox {\boldmath $C$}}
\def\bD{\mbox {\boldmath $D$}}
\def\bE{\mbox {\boldmath $E$}}
\def\bF{\mbox {\boldmath $F$}}
\def\bG{\mbox {\boldmath $G$}}
\def\bH{\mbox {\boldmath $H$}}
\def\bI{\mbox {\boldmath $I$}}
\def\bJ{\mbox {\boldmath $J$}}
\def\bK{\mbox {\boldmath $K$}}
\def\bL{\mbox {\boldmath $L$}}
\def\bM{\mbox {\boldmath $M$}}
\def\bN{\mbox {\boldmath $N$}}
\def\bO{\mbox {\boldmath $O$}}
\def\bP{\mbox {\boldmath $P$}}
\def\bQ{\mbox {\boldmath $Q$}}
\def\bR{\mbox {\boldmath $R$}}
\def\bS{\mbox {\boldmath $S$}}
\def\bT{\mbox {\boldmath $T$}}
\def\bU{\mbox {\boldmath $U$}}
\def\bV{\mbox {\boldmath $V$}}
\def\bW{\mbox {\boldmath $W$}}
\def\bX{\mbox {\boldmath $X$}}
\def\bY{\mbox {\boldmath $Y$}}
\def\bZ{\mbox {\boldmath $Z$}}

\def\ba{\mbox {$\bf{a}$}}
\def\bb{\mbox {\boldmath $b$}}
\def\bc{\mbox {\boldmath $c$}}
\def\bd{\mbox {\boldmath $d$}}
\def\be{\mbox {\boldmath $e$}}
\def\bg{\mbox {\boldmath $g$}}
\def\bh{\mbox {\boldmath $h$}}
\def\bi{\mbox {\boldmath $i$}}
\def\bj{\mbox {\boldmath $j$}}
\def\bk{\mbox {\boldmath $k$}}
\def\bl{\mbox {\boldmath $l$}}
\def\bm{\mbox {\boldmath $m$}}
\def\bn{\mbox {\boldmath $n$}}
\def\bo{\mbox {\boldmath $o$}}
\def\bp{\mbox {\boldmath $p$}}
\def\bq{\mbox {\boldmath $q$}}
\def\br{\mbox {\boldmath $r$}}
\def\bs{\mbox {\boldmath $s$}}
\def\bt{\mbox {\boldmath $t$}}
\def\bu{\mbox {\boldmath $u$}}
\def\bv{\mbox {\boldmath $v$}}
\def\bw{\mbox {\boldmath $w$}}
\def\bx{\mbox {\boldmath $x$}}
\def\by{\mbox {\boldmath $y$}}
\def\bz{\mbox {\boldmath $z$}}

\newcommand{\snr}{\textup{SNR}}
\newcommand{\UE}{\mathrm{UE}}
\newcommand{\BS}{\mathrm{BS}}
\newcommand{\Passoc}{p_{_{I^{(1)}}}}
\newcommand{\Pintra}{p_{_{I^{(2)}}}}
\newcommand{\Pinter}{p_{_{I^{(3)}}}}

\newenvironment{Ex}
{\begin{adjustwidth}{0.04\linewidth}{0cm}
\begingroup\small
\vspace{-1.0em}
\raisebox{-.2em}{\rule{\linewidth}{0.3pt}}
\begin{example}
}
{
\end{example}
\vspace{-5mm}
\rule{\linewidth}{0.3pt}
\endgroup
\end{adjustwidth}}

\newcommand{\Hossein}[1]{{\textcolor{blue}{\emph{**Hossein: #1**}}}}
\newcommand{\Gabor}[1]{{\textcolor{cyan}{\emph{**Afsaneh: #1**}}}}
\newcommand{\Hadi}[1]{{\textcolor{red}{#1}}}
\newcommand{\gf}[1]{{\textcolor{cyan}{#1}}}
\newcommand{\REV}[1]{{\textcolor{blue}{#1}}}


\makeatletter
\let\save@mathaccent\mathaccent
\newcommand*\if@single[3]{%
  \setbox0\hbox{${\mathaccent"0362{#1}}^H$}%
  \setbox2\hbox{${\mathaccent"0362{\kern0pt#1}}^H$}%
  \ifdim\ht0=\ht2 #3\else #2\fi
  }
\newcommand*\rel@kern[1]{\kern#1\dimexpr\macc@kerna}
\newcommand*\widebar[1]{\@ifnextchar^{{\wide@bar{#1}{0}}}{\wide@bar{#1}{1}}}
\newcommand*\wide@bar[2]{\if@single{#1}{\wide@bar@{#1}{#2}{1}}{\wide@bar@{#1}{#2}{2}}}
\newcommand*\wide@bar@[3]{%
  \begingroup
  \def\mathaccent##1##2{%
    \let\mathaccent\save@mathaccent
    \if#32 \let\macc@nucleus\first@char \fi
    \setbox\z@\hbox{$\macc@style{\macc@nucleus}_{}$}%
    \setbox\tw@\hbox{$\macc@style{\macc@nucleus}{}_{}$}%
    \dimen@\wd\tw@
    \advance\dimen@-\wd\z@
    \divide\dimen@ 3
    \@tempdima\wd\tw@
    \advance\@tempdima-\scriptspace
    \divide\@tempdima 10
    \advance\dimen@-\@tempdima
    \ifdim\dimen@>\z@ \dimen@0pt\fi
    \rel@kern{0.6}\kern-\dimen@
    \if#31
      \overline{\rel@kern{-0.6}\kern\dimen@\macc@nucleus\rel@kern{0.4}\kern\dimen@}%
      \advance\dimen@0.4\dimexpr\macc@kerna
      \let\final@kern#2%
      \ifdim\dimen@<\z@ \let\final@kern1\fi
      \if\final@kern1 \kern-\dimen@\fi
    \else
      \overline{\rel@kern{-0.6}\kern\dimen@#1}%
    \fi
  }%
  \macc@depth\@ne
  \let\math@bgroup\@empty \let\math@egroup\macc@set@skewchar
  \mathsurround\z@ \frozen@everymath{\mathgroup\macc@group\relax}%
  \macc@set@skewchar\relax
  \let\mathaccentV\macc@nested@a
  \if#31
    \macc@nested@a\relax111{#1}%
  \else
    \def\gobble@till@marker##1\endmarker{}%
    \futurelet\first@char\gobble@till@marker#1\endmarker
    \ifcat\noexpand\first@char A\else
      \def\first@char{}%
    \fi
    \macc@nested@a\relax111{\first@char}%
  \fi
  \endgroup
}
\makeatother

\def\herm{\mathsf{H}}
\def\trans{\mathsf{T}}
\newcommand{\call}[1]{{\textsf{\small \textsc{#1}}}}
\newcommand{\callf}[1]{{\textsf{\footnotesize \textsc{#1}}}}

\def\argmax{\mathrm{arg}\max}
\def\argmin{\mathrm{arg}\min}
\renewcommand{\algorithmicrequire}{\textbf{Input:}}
\renewcommand{\algorithmicensure}{\textbf{Output:}}
\algdef{SE}[PROCEDURE]{Procedure}{EndProcedure}%
   [2]{\algorithmicprocedure\ \textproc{#1}\ifthenelse{\equal{#2}{}}{}{(#2)}}%
   {\algorithmicend\ \algorithmicprocedure}%
\algdef{SE}[FUNCTION]{Function}{EndFunction}%
   [2]{\algorithmicfunction\ \textproc{#1}\ifthenelse{\equal{#2}{}}{}{(#2)}}%
   {\algorithmicend\ \algorithmicfunction}%

\makeatletter
\newcommand\fs@betterruled{%
  \def\@fs@cfont{\bfseries}\let\@fs@capt\floatc@ruled
  \def\@fs@pre{\vspace*{5pt}\hrule height.8pt depth0pt \kern2pt}%
  \def\@fs@post{\kern2pt\hrule\relax}%
  \def\@fs@mid{\kern2pt\hrule\kern2pt}%
  \let\@fs@iftopcapt\iftrue}
\floatstyle{betterruled}
\restylefloat{algorithm}
\makeatother

\maketitle

\begin{abstract}

Federated learning (FL) is a distributed learning paradigm wherein users exchange FL models with a server instead of raw datasets, thereby preserving data privacy and reducing communication overhead. However, the increased number of FL users may hinder completing large-scale FL over wireless networks due to high imposed latency. Cell-free massive multiple-input multiple-output~(CFmMIMO) is a promising architecture for implementing FL because it serves many users on the same time/frequency resources. While CFmMIMO enhances energy efficiency through spatial multiplexing and collaborative beamforming, it remains crucial to meticulously allocate uplink transmission powers to the FL users. In this paper, we propose an uplink power allocation scheme in FL over CFmMIMO by considering the effect of each user's power on the energy and latency of other users to jointly minimize the users' uplink energy and the latency of FL training. The proposed solution algorithm is based on the coordinate gradient descent method. Numerical results show that our proposed method outperforms the well-known max-sum rate by increasing up to~$27$\% and max-min energy efficiency of the Dinkelbach method by increasing up to~$21$\% in terms of test accuracy while having limited uplink energy and latency budget for FL over CFmMIMO. 
\end{abstract}
\section{Introduction}

Federated learning~(FL) is an emerging paradigm allowing users to jointly train a machine learning algorithm without sharing their local dataset~\cite{konevcny2016federated}. In each iteration of the FL training, each user trains a local copy of the algorithm using its private dataset; then, the users transmit their local FL model updates to a server that uses them to update the global model parameter and broadcast it to the users. Transmitting the local model updates instead of the raw datasets significantly reduces the communication overhead and preserves the users' data privacy. In practice, an efficient communication network is required to facilitate a stable and fast exchange of the high-dimensional local models and thereby achieve successful FL. Cell-free massive multiple-input multiple-output~(CFmMIMO) is a suitable infrastructure for FL since it is designed to provide nearly uniform quality-of-service (QoS) \cite{cellfreebook}, as required when all users have equal-sized model updates to upload. CFmMIMO networks consist of a large number of spatially distributed access points (APs) that serve the users through coherent joint transmission and reception, thereby harnessing macro-diversity and channel hardening, which makes the data rates predictable even with imperfect channel knowledge.

While FL reduces the data communication overhead, the users' energy limitations may hinder completing the high-dimensional, large-scale FL training, particularly when it comes to transmissions of local models~\cite{9311931}. Therefore, efficient resource allocation (e.g., time, space, or energy) is crucial to ensuring the successful execution of FL training. In this regard, \cite{9796621,9975256,9500541,9124715,8781848,vu2021does,10120750} have proposed approaches to efficiently allocate resources for FL training in CFmMIMO. This combination is attractive since the model updates are equally sized for all users, and CFmMIMO can deliver nearly uniform QoS.

To reduce the latency in FL training, \cite{9796621} has considered a joint optimization problem of user selection, transmit power, and processing frequency to mitigate the high latency of users with unfavorable communication links. Furthermore, the authors in~\cite{9500541,9975256} have proposed jointly optimizing the transmit power and data rate to reduce the uplink latency of a CFmMIMO-supported-FL system. The reference~\cite{9124715} has proposed a joint optimization of the local accuracy, transmit power, data rate, and users' processing frequency to minimize the latency of FL in CFmMIMO. Furthermore, \cite{vu2021does} has considered a CFmMIMO scheme to support multiple FL groups and proposed an algorithm to allocate power and processing frequency to reduce the latency of each FL iteration optimally. 

Energy-efficient FL over CFmMIMO has recently attracted the attention of several researchers aiming to achieve successful FL training in energy-constrained scenarios. Energy efficiency maximization was considered in \cite{8781848} with per-user power, backhaul capacity, and throughput constraints. The authors derived a closed-form expression for spectral efficiency, considering the effects of channel estimation error and quantization distortion. In~\cite{10120750}, an energy-efficient FL scheme over cell-free internet-of-things networks was designed to minimize the total energy consumption of the FL users. The authors obtained an optimal central processing unit and a sub-optimal power allocation solution for their proposed optimization problem. 

The papers mentioned above on FL over CFmMIMO networks either consider energy or latency optimization.
However, in practice, it is necessary to strike a balance between these conflicting goals. The trade-off is non-trivial since the lowest energy is achieved by transmitting at very low rates, which leads to high latency. Additionally, the effect of users' interference on each user should be considered while jointly optimizing the uplink energy and latency.  

 In this paper, we aim to allocate the uplink powers optimally to the FL users in a CFmMIMO network to jointly minimize each user's communication energy and latency. To this aim, we propose to minimize a weighted sum of the users' uplink energy and latency. The proposed trade-off metric models two important factors:~($\boldsymbol{i}$) minimizing the trade-off between each user's uplink energy and latency, and~($\boldsymbol{ii}$) allocating the uplink powers by considering the effect of each user's power on the energy and latency of other users while jointly optimizing the uplink energy and latency. Considering factor~($\boldsymbol{ii}$) is crucial because FL training requires the simultaneous participation of many FL users in a network. Thus, considering the impacts of the users on each other is necessary for successful and efficient FL training. We show that the objective of the optimization problem {\color{black}has a unique minimum w.r.t. each FL user's uplink power}, and we derive a solution algorithm based on the coordinate gradient descent method. We compare the final test accuracy of applying our approach with two well-known benchmarks of max-sum rate~\cite{cellfreebook} and max-min energy-efficiency generalized Dinkelbach~\cite{EE_dinkel} methods (for simplicity, we call it Dinkelbach through the paper) in a limited energy and latency budget condition in FL over CFmMIMO. The numerical results show that while max-sum rate and Dinkelbach methods are not optimal in this regard, our proposed approach obtains the optimal power allocation policy that outperforms max-sum rate and Dinkelbach by up {to~$27$\% and~$21$\%} increase in the FL test accuracy. To the best of our knowledge, this is the first work that considers jointly minimizing uplink energy and latency in FL over CFmMIMO.    

The paper is organized as follows. Section~\ref{section: System Model and Problem Formulation} describes the system models of FL and CFmMIMO and the problem formulation. In Section~III, we derive an iterative algorithm for solving the proposed optimization problem with closed-form updates. Section~IV provides numerical results to demonstrate the benefits of the algorithm compared to Dinkelbach and max-sum methods, and the conclusions are stated in Section~V.

\emph{Notation:} Italic font $w$, boldface lower-case $\bw$, boldface upper-case $\bW$, and calligraphic font $\calW$ denote scalars, vectors, matrices, and sets, respectively. We define the index set $[N] = \{1,2,\ldots,N\}$ for any positive integer $N$. We denote the $l_2$-norm by $\|\cdot\|$ , the floor value by $\floor{\cdot}$, the cardinality of set $\calA$ by $|\calA|$, the entry $i$ of the vector $\bw$ by $[\bw]_i$, the entry $i,j$ of the matrix $\bW$ by $[w]_{i,j}$, and the transpose of $\bw$ by $\bw{\tran}$. 
\begin{figure}[t]
\centering
 \includegraphics[width=0.91\columnwidth]{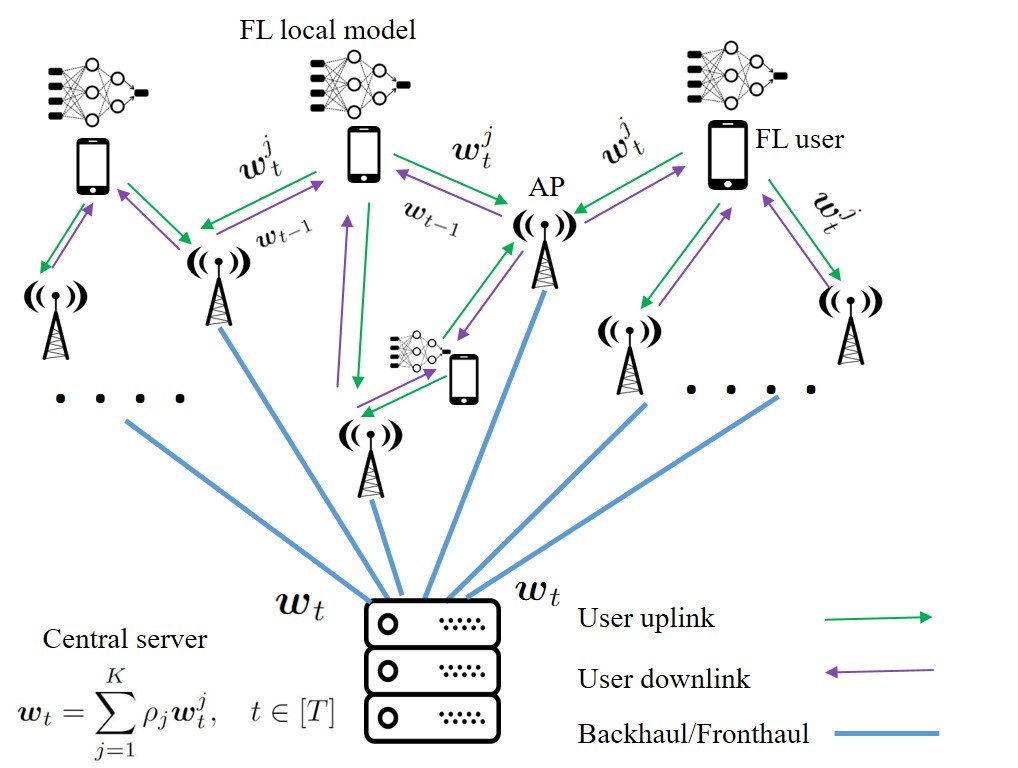}
  
  \caption{General architecture of FL over CFmMIMO. }\label{fig: CFmMIMO_FL}
\end{figure}
\section{System Model and Problem Formulation}\label{section: System Model and Problem Formulation}
This section describes the FL setup, presents the system model, and states the problem formulation. 

\subsection{Federated Averaging}\label{subsec: FLs}
We consider a setup where $K$ FL users/clients cooperatively solve a distributed learning problem involving a loss function $f(\bw)$. The dataset $\calD$ is distributed among the users. The disjoint subset
$\calD_j$ is available at user $j \in [K]$, which satisfies $\calD_j \cap \calD_{j'}=\emptyset$ for $j \neq j'$ and $\sum_{j =1}^{K} |\calD_j| = |\calD|$.
We let the tuple $(\bx_{ij}, y_{ij})$ denote data sample $i$ of the $|\calD_j|$ samples available at user $j$.
We let $\bw \in \R^d$ denote the global model parameter with dimension~$d$ and define $\rho_j := {|\calD_j|}/{|\calD|}$ as the fraction of data available at user $j$. We formulate the training problem
\vspace{-0.005\textheight}
\begin{equation}\label{eq: w*}
\bw^* \in \argmin_{\bw\in\R^d} f(\bw)=\sum_{j =1}^{K} 
{\rho_j f_j(\bw)},
\vspace{-0.008\textheight}
\end{equation}
where $f_j(\bw) := \sum_{i =1}^{|\calD_j|} {f(\bw; \bx_{ij}, y_{ij})}/{|\calD_j|}$. To solve~\eqref{eq: w*}, we use federated averaging~(FedAvg)~\cite{li2019convergence}, an iterative algorithm with $T$ iterations. Initializing the training with~$\bw_0$, at the beginning of each iteration~$t \in [T]$, the server sends $\bw_{t-1}$ to the users. Then, every user $j\in[K]$ performs $L$ local iterations, $i=1, \ldots,l$, of stochastic gradient descent~\cite{8889996} with randomly-chosen subset of $\xi_j \le |\calD_j|$ data samples, and computes its local model~$\bw_{i,t}^j~\in~\R^d$, considering the initial point of $\bw_{0,t}^{j} = \bw_{t-1}$:
\vspace{-0.015\textheight}
\begin{equation}\label{eq: local trains}
  \bw_{i,t}^{j} \xleftarrow{} \bw_{i-1,t}^{j}-\frac{\alpha_t}{\xi_j} \sum_{n=1}^{\xi_j} \nabla_{\bw} f(\bw_{i-1,t}^{j}; \bx_{nj}, y_{nj}), \,\, t\in[T],
  \vspace{-0.008\textheight}
 \end{equation}
 where~$\alpha_t$ is the step size and the final local model is $\bw_t^j = \bw_{L,t}^{j}$. After computing~$\bw_t^j$ at iteration $t$, user $j$ transmits it to the central server over a wireless interface. More precisely, we consider a CFmMIMO system with $M$ APs, which forward their received signals to the server over error-free fronthaul links, as shown in Fig.~\ref{fig: CFmMIMO_FL}. Based on the $K$ local models,
 the central server updates the global model~$\bw_{t}$ by performing a weighted average of the local models~$\bw_t^1,\ldots,\bw_t^K$ as 
  \vspace{-0.01\textheight}
 \begin{equation}\label{eq: iterative}
    \bw_{t} = \sum_{j =1}^{K}  \rho_j \bw_t^j.
     \vspace{-0.01\textheight}
\end{equation}
Afterward, the central server sends the global model~$\bw_{t}$ to 
the APs via the fronthaul links. Next, the APs jointly send~$\bw_{t}$ to the users in the downlink. Finally, each user~$j \in [K]$ computes its new local model at the beginning of the next FL iteration. 

In this paper, we focus on the uplink transmission for sending the local models to the APs, with the goal of making it as resource-efficient as possible. We stress that the uplink is the weakest link since there are $K$ local models to transfer with limited energy from battery-powered devices, while there is only one global model to transmit with high power from grid-connected APs in the downlink. Thus, without loss of generality, we assume that the downlink is error-free and all users receive the same~$\bw_t$. 

\vspace{-0.002\textheight}
\subsection{Uplink Process in CFmMIMO}
We consider a CFmMIMO system consisting of $M$ APs, each with $N$ antennas and $K$ single-antenna FL users that will transmit their local models. We consider the standard block-fading model where each channel takes one realization per time-frequency coherence block of~$\tau_c$ channel uses and independent realizations across blocks~\cite{cellfreebook}. 
We assume independent Rayleigh fading where $g_{m,n}^{j} \sim \mathcal{N}_{\mathbb{C}}(0, \beta_{m}^{j})$ is the channel coefficient between user~$j$ and the~$n${{th}} antenna of AP~$m$, 
where~$\beta_{m}^{j}$ is the large-scale fading coefficient.
We focus on the uplink where the coherence block length $\tau_c$ is divided into $\tau_p < \tau_c$ pilot channel uses, and $\tau_c-\tau_p$ data channel uses.

\subsubsection{Channel Estimation}

Each user is assigned a $\tau_p$-length pilot from a set of $\tau_p$ mutually orthogonal pilot sequences. 
The pilot of user $j$ is denoted $\sqrt{\tau_p} \boldsymbol{\varphi}_j~\in~\mathbb{C}^{\tau_p\times 1}$, where $\|\boldsymbol{\varphi}_j\|^2 = 1$.
The users send these sequences simultaneously and then the received vector~$\by_{m,n}^{p} \in \mathbb{C}^{\tau_p \times 1}$ at antenna~$n$ of the AP $m$ is
\vspace{-0.017\textheight}
\begin{equation}\label{eq: ypnl}
  \by_{m,n}^{p} = \sqrt{p_p}\sum_{j=1}^K g_{m,n}^{j} \boldsymbol{\varphi}_j + \boldsymbol{\mu}_{m,n}^{p}, 
  \vspace{-0.0085\textheight}
\end{equation}
where $p_p := \tau_p \rho_p$ is total normalized energy of the pilot transmission with~$\rho_p$ being the normalized signal-to-noise ratio (SNR) per pilot, and~$\boldsymbol{\mu}_{m,n}^{p}$ is the noise vector with i.i.d. $\mathcal{N}_{\mathbb{C}} (0, 1)$-entries. By projecting~$\by_{m,n}^{p}$ onto the pilot~$\boldsymbol{\varphi}_j$, we obtain~$\hat{y}_{m,n}^{p,j} = \boldsymbol{\varphi}_j^H \by_{m,n}^{p} $. Then, the MMSE estimate of~$g_{m,n}^{j}$ is obtained as~\cite{cellfreebook}
\vspace{-0.008\textheight}
\begin{equation}\label{eq: ghat}
   \hat{g}_{m,n}^{j} = \frac{\mathds{E}\left\{ {\hat{y}_{m,n}^{p,j}}{g^*}_{m,n}^{j}\right\}}{\mathds{E}\left\{ \left|\hat{y}_{m,n}^{p,j}\right|^2\right\}} = c_{m}^{j} \hat{y}_{m,n}^{p,j},
\end{equation}
where
\vspace{-0.015\textheight}
\begin{equation}\label{eq: cnlk}
    c_{m}^{j} := \frac{\sqrt{p_p} \beta_{m}^{j}}{p_p \sum_{j'=1}^{K} \beta_{m}^{j'} | \boldsymbol{\varphi}_{j'}^H \boldsymbol{\varphi}_j |^2 + 1}. 
    \vspace{-0.008\textheight}
\end{equation}
\vspace{-0.002\textheight}
\subsubsection{Uplink Data Transmission}
We consider that all the $K$ users simultaneously send their local models to the APs in the uplink. We define~$q_j$, $j=1,\ldots,K$, as the independent unit-power data symbol associated with the $j$th user (i.e., $\mathds{E}\left\{ |q_j|^2 \right\} = 1$). User~$j$ sends its signal using the power~{\color{black}$p^u p_j$}, where $0 \le p_j \le 1$ is the power control coefficient we will optimize later. 
Then, the received signal~$ y_{m,n}^{u,j}$ at antenna~$n$ of AP~$m$ is
\vspace{-0.016\textheight}
\begin{equation}\label{eq: ymnk}
    y_{m,n}^{u} = \sum_{j=1}^K g_{m,n}^{j} q_j \sqrt{{\color{black}{p^u}} p_j} + \mu_{m,n}^u,
    \vspace{-0.0048\textheight}
\end{equation}
where~$\mu_{m,n}^u \sim \mathcal{N}_{\mathbb{C}}(0,{\color{black}\sigma^2})$ is the additive
noise at the $n$th antenna of AP~$m$. After the APs receive the uplink signal, they use maximum-ratio processing. Similar to~\cite[Th.~2]{7827017}, the achievable uplink data rate (in bit/s) at user $j$ is
\vspace{-0.0135\textheight}
\begin{alignat}{3}\label{eq: Rk}
    &R_j = \: \\
    \nonumber
    \vspace{-0.018\textheight}
     &\left( 1 - \frac{ \tau_p}{\tau_c} \right) B \log_2\left(1+ \frac{p_j \left( \sum\limits_{m=1}^M N{\gamma_m^j}\right)^2}{ \sum\limits_{j' = 1}^K p_{j'} \sum\limits_{m=1}^M N \gamma_m^j \beta_{m}^{j'} + I_p^{j} + I_M^{j}}\right),
\end{alignat}
where~${\gamma_{m}^{j}} := \mathds{E}\left\{ |\hat{g}_{m,n}^j|^2\right\} = \sqrt{p_p} \beta_{m}^j c_{m}^j $, $B$ is the bandwidth~(Hz), and
 \vspace{-0.015\textheight}
\begin{alignat}{3}
\label{eq: IM and Ip}    
I_M^{j} &= \sum_{m=1}^M~N{\color{black}\sigma^2{\gamma_{m}^{j}}}/{{p^u}}, \: \\
\nonumber
I_p^{j} &= \sum_{\substack{{j'}=1\\j' \neq j}}^{K} p_{j'} |\boldsymbol{\varphi}_j^{H} \boldsymbol{\varphi}_{j'}|^2\left(\sum_{m=1}^M N {\gamma_{m}^{j}}\frac{\beta_{m}^{j'}}{\beta_{m}^{j}}\right)^2,
\end{alignat}
where~$\sigma^2$ is the uplink noise power, and~$p^u$ is the maximum uplink transmit power.
\vspace{-0.005\textheight}
\subsection{Problem Formulation}
\vspace{-0.005\textheight}
 Here, we state the main problem formulation of this paper, which aims at training an FL algorithm under uplink energy and latency constraints by optimizing the uplink power of each user~$p_j$, $j = 1, \ldots, K$. To this end, we define the uplink latency as~$\ell_j:=bd/R_j$ and the uplink energy~$E_j$ of user~$j$ as
 \vspace{-0.012\textheight}
 \begin{equation}\label{eq: energy}
     E_j := {\color{black}{p^u}} p_j \ell_j = {\color{black}{p^u}} p_j\frac{b d }{R_j} ,\quad j=1,\ldots, K,
 \end{equation}
where~$b$ is the number of bits required to send each entry of $\bw_t^j~\in~\R^d$ and $d$ is the size of the local and global FL models. We define $\bp:= [p_1,\ldots, p_K]^{\intercal} \in \R^{K \times 1}$ as the vectorized user uplink power coefficients. We propose the following optimization problem that minimizes the weighted sum of the uplink energy and uplink latency of users as
\vspace{-0.01\textheight}
\begin{subequations}\label{eq: main optimization}
\begin{alignat}{3}
\label{optimization1}
  \underset{p_1, \ldots, p_K}{\mathrm{minimize}} & \quad \sum_{j=1}^K \theta_1 E_j + \theta_2 \ell_j \: \\ 
  \text{subject to}
  \label{constraint1}
 &\quad 0 \le p_j \le 1,\quad j=1,\ldots,K,
\end{alignat}
\end{subequations}
where~$\theta_1, \theta_2~\in [0,1]$ are the weights that can compensate for any scaling differences between the energy and latency. We note that for every user~$j~\in~[M]$,~$E_j$ is an increasing function of~$p_j$ while~$\ell_j$ is a decreasing function of~$p_j$, when considering~$0\le~p_j\le~1$. As a result, any linear combination of~$\theta_1 E_j + \theta_2 \ell_j$ indicates the trade-off between~$E_j$~and~$\ell_j$~\cite{boydcnvx}, for~$p_j~\in[0,1]$, and the weighted sum in~\eqref{optimization1} denotes the overall trade-off between the uplink energy and latency of the FL users at every iteration~$t~\in~[T]$. Consequently, this overall trade-off demonstrates the effect of each~$p_j$ on the values of~$E_{k'}$ and~$\ell_{k'}$ of other FL users~$k'~\neq j$, which optimization problem~\eqref{eq: main optimization} takes into account for optimal power allocation. We will show the detailed mathematical expressions in Section~III. Therefore, by solving the optimization problem~\eqref{eq: main optimization}, we obtain the uplink powers of each user~$j$ that jointly minimizes the trade-off between energy~$E_j$ and latency~$\ell_j$ for all users~$j=1,\ldots, K$.
\begin{lemma}\label{lemma: J_cnx}
Let~$E_j$ and~$\ell_j$ be the energy and latency for sending each local FL model~$\bw_t^j$ to the APs. We define~${\nu}(\bp;\theta_1,\theta_2, b, d):= \sum_{j=1}^K \theta_1~E_j~+~\theta_2~\ell_j$ as the objective  function in~\eqref{optimization1}. When all other variables are fixed, ${\nu}(\bp;\theta_1, \theta_2, b, d)$ {\color{black}has a unique minimum w.r.t. $p_j$ (for any user $j=1, \ldots, K$)}, defined as
\begin{subequations}\label{eq: main optimization2}
\begin{alignat}{3}
\label{optimization2}
  p_j^*~& \in \quad \underset{{{p_j}}}{\argmin}\quad {\nu}(\bp;\theta_1, \theta_2, b, d) \: \\ 
  \text{subject to}
  \label{constraint2}
 &\quad 0 \le p_j \le 1. 
\end{alignat}
\end{subequations}
\end{lemma}
\begin{IEEEproof}
    The proof is given in Appendix~\ref{P:lemma: J_cnx}.
\end{IEEEproof}
Lemma~\ref{lemma: J_cnx} shows that the objective loss function ${\nu}(\bp; \theta_1, \theta_2,  b, d)$ {\color{black}has a unique minimum w.r.t every $p_j$, $j=1,\ldots, K$.} We will use this property in Section III to devise an iterative solution algorithm. We define~$\mathcal{L}$ and~$\mathcal{E}$ as the total uplink latency and energy budget that all the users can spend for the uplink phase of the FL training. By obtaining the uplink powers, $\bp$, that minimize the latency and energy per iteration, the number of the global FL iterations~$T$ can be maximized under the given energy and latency constraints. Therefore, after solving the problem, we consider~$E_j^*:=E_j|_{p_j = p_j^*}$,~$\ell_j^*:=\ell_j|_{p_j = p_j^*}$ and define~$\bar{\ell}^* := \max_j ~\ell_j^*$ and ~$\bar{E}^* := \sum_{j=1}^K ~E_j^*$, and we obtain the maximum number of FL iterations as
\vspace{-0.005\textheight}
\begin{equation}\label{eq: iterations}
    T_{\max} := \min \left\{ \frac{\mathcal{L}}{ \bar{ \ell}^* }, \frac{\mathcal{E}}{ \bar{E}^*} \right\}.
\end{equation}
As a result, we perform an FL training with the maximum number of global FL iterations~$T = \floor{T_{\max}}$ based on the optimized uplink powers~$p_j^*$. In the following section, we propose an algorithm to solve~\eqref{optimization1} using Lemma~\ref{lemma: J_cnx}. 

\begin{algorithm}[t]
\caption{FL over CFmMIMO with uplink power allocation.} 
\small
\label{alg: FL over CFmMIMO}
\begin{algorithmic}[1]
\State \textbf{Inputs: $M$, $N$, $K$, $d$, $\eta$, $\bar{\epsilon}$, $\varepsilon$, ${(\bx_{ij}, y_{ij})}_{i,j}$, ${\color{black}{\gamma_{m}^{j}}}$, $\beta_{m}^j$, $I_p^{j}$, $I_M^j$, {$\forall j, m, n$}}, $\alpha_t$, $\bw_0$, $\theta_1$, $\theta_2$, $p^u$, $\sigma^2$

\State \textbf{Initialize:} $\bp_{0} = \boldsymbol{0}$, $\bu = \boldsymbol{1}$, $\{p_j\}_{j=1,\ldots, K} = 0$, $t=1$, $\partial {\nu}_j= +\infty$ 
\While{$\|\bu - \bp_{0}\|_{\infty} > \varepsilon$,} \Comment{{\color{cyan}Power allocation}}
\For{$j =1,\ldots,K$,}
\While{$|\partial {\nu}_j| > \bar{\epsilon}$ and $p_j < 1$,} \Comment{{\color{cyan}Coordinate descent}}

\State Set $p_{j,0} = p_j$
\State Compute~$\partial {\nu}_j=\frac{\partial {\nu}_j}{\partial p_j} |_{p_j = p_{j,0}}$ according to~\eqref{eq: p_j^* solution}

\State Update $p_j \leftarrow p_{j,0} -  \eta \partial \nu_j$  

\EndWhile 
\State Set $[\bu]_j = p_j$, $[\bp_0]_j = p_{j,0}$
\EndFor
\EndWhile 
\State $p_j^* = [\bu]_j$, $\bp^* = [p_1^*,\ldots,p_K^*]^{\intercal}$

\State Compute~$T_{\max}$~according to~\eqref{eq: iterations} \Comment{{\color{cyan}FL training}}
\State Set $T \le \floor{T_{\max}}$

\For{$t \le T$,}  \Comment{{\color{cyan}Global iterations}}
\State APs receive~$\bw_{t-1}$ from the central server and send it to all users via downlink transmission
    \For{$j = 1, \ldots, K$,} 
        \State Set $\bw_{0,t}^j=\bw_{t-1} $ 
        \For{$i= 1, \ldots, L$} \Comment{{\color{cyan}Local iterations}}
            \State update $\bw_{i,t}^j$ according to~\eqref{eq: local trains}
            
            
        \EndFor
        \State Set $\bw_{t}^j=\bw_{L,t}^j $ 
        \State Send $\bw_{t}^j$ to the APs via uplink transmission
    \EndFor
\State Central server waits it until receives all $\bw_{t}^j$, $ {j=1,\ldots,K}$ from the APs and updates $\bw_{t} \leftarrow \sum_{j =1}^{ K}  \rho_j \bw_{t}^j$ 
    \EndFor

\State \textbf{Return} $\bp^*$, $\bw_T$

\end{algorithmic}

\end{algorithm}


\section{Solution Algorithm}
This section presents the proposed solution algorithm for optimization problem~\eqref{optimization1}. First, we provide the solution to the single-variable optimization problem~\eqref{eq: main optimization2} to obtain~$p_j^*$, for $j=1,\ldots, K$. 
Afterward, we describe Algorithm~\ref{alg: FL over CFmMIMO} that solves these problems iteratively until convergence. Finally, we utilize~$p_j^*$ to obtain~$T_{\max}$ and perform the FL iterations for $T \le \floor{T_{\max}}$ iterations and obtain the global FL model~$\bw_{T}$.
\begin{prop}\label{prop: obtain p_j^*}
Consider ${\nu}(\bp; \theta_1, \theta_2,  b, d)=\sum_{j=1}^K \theta_1 E_j + \theta_2 \ell_j$. Let~$\bp^* = [p_1^*,\ldots,p_K^*]^{\intercal}$ be the vectorized version of the solutions obtained after solving~\eqref{eq: main optimization2}. We re-write~${\nu}(\bp;\theta_1,\theta_2, b, d)$ as
\vspace{-0.01\textheight}
\begin{alignat}{3}\label{eq: nu definition}
{\nu}&(\bp;\theta_1,\theta_2, b, d) = \frac{\theta_1 {\color{black}{p^u}} p_j + \theta_2}{\log_2\left(1 + p_j \bar{A}_j/\left(p_j\bar{B}_j + \bar{C}_j\right)\right)} + \: 
\nonumber \\
&\sum_{\substack{\substack{{k'}=1\\k' \neq j}}}^K \frac{\theta_1 {\color{black}{p^u}} p_{k'} + \theta_2}{\log_2\left(1 + p_{k'} \bar{A}_{k'}/\left(p_j\Tilde{B}_{k'}^{j} + \Tilde{C}_{k'}^{j}\right) \right)},~\forall j\in[K],
\end{alignat}
which~${\nu}(\bp; \theta_1, \theta_2,  b, d)$ {\color{black}is differentiable and has a unique minimum w.r.t. every~$p_j$}, then~$p_j^*$ is obtained as 
\begin{equation}\label{eq: p_j^* define}
    p_j^* = \left\{ p_j :~\partial {\nu}_j := \frac{\partial }{\partial p_j}{\nu}(\bp; \theta_1, \theta_2,  b, d) = 0 \right\}, 
\end{equation}
where
  \begin{align}\label{eq: p_j^* solution}
     \partial {\nu}_j &= \frac{ \theta_1 {\color{black}{p^u}} \log_2(1 + p_j \bar{A}_j/\left(p_j\bar{B}_j + \bar{C}_j)\right) - (\theta_2 +\theta_1 {\color{black}{p^u}} p_j){\bar{L}_{p_j}}}{\left(\log_2\left(1 + p_j \bar{A}_j/\left(p_j\bar{B}_j + \bar{C}_j\right)\right)\right)^2}\:
     \nonumber \\
      & - \sum_{\substack{\substack{{k'}=1\\k' \neq j}}}^K \frac{(\theta_2 +\theta_1 {\color{black}{p^u}} p_{k'})\Tilde{L}_{k'}^{j}}{\left(\log_2\left(1 + p_{k'} \bar{A}_{k'}/\left(p_j\Tilde{B}_{k'}^{j} + \Tilde{C}_{k'}^{j}\right) \right)\right)^2},
  \end{align}

\vspace{-0.01\textheight}
  \begin{alignat}{3}\label{eq: definitions}
      \bar{A}_j &:= \left( \sum_{m=1}^M N{\color{black}{\gamma_{m}^{j}}}\right)^2,\quad \bar{B}_j := \sum_{m=1}^M N {\gamma_{m}^{j}} \beta_{m}^{j}, \: \\
      \bar{C}_j &:= I_p^{j} + I_M^{j} + \sum_{{j'} \neq j}^{K} {p}_{j'} \sum_{m=1}^M N {\color{black}{\gamma_{m}^{j}}} \beta_{m}^{j'},\: \\
      \bar{L}_{p_j} &:= \frac{\bar{A}_j \bar{C}_j}{((\bar{A}_j +\bar{B}_j )p_j + \bar{C}_j )(\bar{B}_j p_j + \bar{C}_j )\ln(2)},\: \\
      \Tilde{B}_{k'}^{j} &:= \sum_{m=1}^M N {\gamma_{m}^{k'}} \beta_{m}^{j} + |\boldsymbol{\varphi}_{k'}^{H} \boldsymbol{\varphi}_{j}|^2\left(\sum_{m=1}^M N {\gamma_{m}^{k'}}\frac{\beta_{m}^{j}}{\beta_{m}^{k'}}\right)^2,  \: \\
      \Tilde{C}_{k'}^{j} &:= \bar{B}_{k'} p_{k'} + I_M^{k'} + I_p^{k'} - \Tilde{B}_{k'}^{j} p_j+\sum_{\substack{{j'}=1\\j' \neq j, k'}}^{K}\sum_{m=1}^M p_{j'}N {\gamma_{m}^{k'}} \beta_{m}^{j'}\: \\
        \label{eq: definitions1}
      \Tilde{L}_{k'}^{j} &:= \frac{-\Tilde{B}_{k'}^{j} \bar{A}_{k'} p_{k'} }{(\Tilde{B}_{k'}^{j} p_j + \Tilde{C}_{k'}^{j} )(\Tilde{B}_{k'}^{j} p_j + \bar{A}_{k'} p_{k'} + \Tilde{C}_{k'}^{j})\ln(2)}.
      \end{alignat}
\end{prop}
Proposition~\ref{prop: obtain p_j^*} proposes an approach for solving the single-variable optimization problem~\eqref{eq: main optimization2} when the other power variables are fixed. Since their solutions are interdependent, we propose the iterative method presented in Algorithm~\ref{alg: FL over CFmMIMO} for FL over CFmMIMO with our proposed power allocation scheme. We use the coordinated gradient descent method for our proposed power allocation scheme because it greatly reduces the computation overhead of solving the optimization problem~\eqref{eq: main optimization}. We note that the second term in~\eqref{eq: p_j^* solution} considers~$\Tilde{B}_{k'}^{j}, \Tilde{C}_{k'}^{j}, \Tilde{L}_{k'}^{j}$ which are the parameters showing how the power~$p_j$ of user~$j$ impacts on the values of energy and latency of other users~$k'~\neq j$. The power allocation part in Algorithm~\ref{alg: FL over CFmMIMO}~(lines~3-13) deploys iterative coordinate descent, such that each user~$j=1,\ldots, K$ performs gradient descent based on the derivative in \eqref{eq: p_j^* define} to obtain~$p_j^*$ while assuming~$p_{j'}, j~\neq~j'$ is constant. Since the objective function is improved in each outer iteration, it converges to the final solution. After computing~$p_j^*$, Algorithm~\ref{alg: FL over CFmMIMO} computes~$T_{\max}$, as in Eq.~\eqref{eq: iterations}, and follows the FL training.





\section{Numerical Results}
We consider a CFmMIMO network with $M=16$ APs, each with~$N=4$ antennas, a central server, and $K \in \{ 20,40\}$ single-antenna FL users. Each user is assigned to a $\tau_p$-length pilot sequence according to the pilot assignment algorithm in~\cite{cellfreebook}. The simulation parameters and values are shown in TABLE~\ref{tab: parameters}. The normalized SNR is computed by dividing the uplink transmit power by the noise power. We consider the CIFAR-10 dataset distributed among the users. 
 The users train Convolutional Neural Networks (CNNs) on their local datasets. The CNNs start with a Conv2D layer (32 filters, (3, 3) size, ReLU activation) processing input data of shape (32, 32, 3). A MaxPooling2D layer (pool size 2x2) follows to downsample spatial dimensions. A Flatten layer converts data to 1D for fully connected layers. Two Dense layers have 64 units (ReLU activation) and 10 units (softmax activation) for multiclass classification. The final layer outputs probabilities for 10 CIFAR-10 classes. The CNN has $d =462,410$ trainable parameters, which is the same for all users. This architecture captures spatial features and classifies them through a softmax layer. After computing each local model, the FL users transmit them simultaneously via uplink communication to the APs. 


\begin{table}[t]
    \centering
    \caption{Simulation parameters for FL over CFmMIMO.}
    \label{tab: parameters}
    \renewcommand{\arraystretch}{1.25} 
    \setlength{\extrarowheight}{1.25pt} 
    
    \resizebox{\columnwidth}{!}{%
    \selectfont
    \begin{tabular}{|l l||l l|}
        \hline
        \textbf{Parameter} &  \textbf{Value} & \textbf{Parameter} &  \textbf{Value}\\
        \hline
        Bandwidth & $B = 20$\,MHz & Number of users & {$K \in \{20, 40\}$}\\
        \hline
        Area of interest (wrap around) & $1000 \times 1000$m & Pathloss exponent & {$\alpha_p= 3.67$}\\
        \hline
        Number of APs & $M = 16$ & Coherence block length & {$\tau_c = 200$}\\
        \hline
        Number of per-AP antennas  & $N = 4$ & Pilot length & {$\tau_p = 10$}\\
        \hline
        Uplink transmit power & {$p^u = 100$\,mW} & Uplink noise power & {$\sigma^2 = -94$\,dBm}\\
        \hline
        Noise figure & {$7$\,dB} & Size of FL model & { {$d = 462410$ }}\\
        \hline
        Number of FL local iterations & { {$L \in \{2, 5\}$ }} & Number of bits & { {$b = 32$ }}\\
        \hline
    \end{tabular}%
    }
\end{table}
 \begin{table}[t]
    \centering
    \caption{Comparison of the final test accuracy of our approach and the benchmarks, with CIFAR-10 dataset,~$\theta_1 =1$, and~$\theta_2=1$. }
    \label{tab: comparison}
    \renewcommand{\arraystretch}{1.25} 
    \setlength{\extrarowheight}{1.5pt} 
   
    \resizebox{\columnwidth}{!}{%
    \fontsize{13}{12}
    
\begin{tabular}{|c|| c|c|c|| c|c|c|}

\hline
 &\multicolumn{3}{c||}{$K = 20$,~$\mathcal{L} = 0.2$,~ $\mathcal{E}=0.2$}&\multicolumn{3}{c|}{$K = 40$,~$\mathcal{L} = 1$,~ $\mathcal{E}=0.5$}\\
\cline{2-7}
&$T = $ & {\large Accuracy}& {\large Accuracy} & $T = $ &{\large Accuracy} & {\large Accuracy}\\
&$\lfloor T_{\max} \rfloor$ & $L = 5$ & $L = 2$  & $\lfloor T_{\max} \rfloor$ & $L = 5$ & $L = 2$\\
\hline
{\textbf{Our approach}} &$75$ &$55.5\%$ & $47.5\%$ &$78$ & $58\%$& $45\%$  \\
\hline
{\textbf{Dinkelbach}} &$42$ & $45\%$ & $38\%$ & $22$& $37\%$ & $27\%$ \\
\hline
{\textbf{Max-sum rate}} &$15$ &$35.5\%$ & $30.8\%$ & $12$ & $31\%$  & $21\%$   \\
\hline
\end{tabular}
 }  
\end{table}
 
 \begin{figure}[t]
\centering
\begin{minipage}{0.6\columnwidth}
{\scriptsize\begin{tikzpicture}
\definecolor{amber}{rgb}{1.0, 0.49, 0.0}
\definecolor{taupe}{rgb}{0.28, 0.24, 0.2}
\definecolor{tealgreen}{rgb}{0.0, 0.51, 0.5}
\definecolor{airforceblue}{rgb}{0.36, 0.54, 0.66} 
\definecolor{darkcandyapplered}{rgb}{0.64, 0.0, 0.0}
\definecolor{columbiablue}{rgb}{0.61, 0.87, 1.0}
\definecolor{darkcandyapplered}{rgb}{0.64, 0.0, 0.0}
\definecolor{oxfordblue}{rgb}{0.0, 0.13, 0.28}
\begin{axis}[%
width=0.99\columnwidth,
height=0.4\columnwidth,
at={(0,0)},
scale only axis,
xmin=-1, 
xmax=95,
xlabel={Global iteration},
xtick= {1, 15, 30, 50, 65, 75, 87},
xticklabels={ {\textbf{1}},15,30, {\textbf{50}}, 65, {\textbf{75}}, {\textbf{87}}},
xtick style={color=black},
ymin=-3,
ymax=60,
ytick={10, 20,30,40,50,60},
grid style={dashed},
ymajorgrids,
grid=both,
ylabel near ticks,
ylabel={Test accuracy (\%)},
axis background/.style={fill=white},
legend style={at={(1,0.02)}, anchor=south east, legend cell align=left,inner sep=0.5pt, font = \tiny, fill=none, text height=0.4ex, text depth=0.3ex}
]

\addplot [color=darkcandyapplered, densely dotted, line width = 0.8pt]
 table[row sep=crcr]{%
1	10.4500003159046\\
2	20.8700001239777\\
3	23.2500001788139\\
4	25.5899995565414\\
5	29.5300006866455\\
6	32.8599989414215\\
7	33.7900012731552\\
8	34.9400013685226\\
9	35.6499999761581\\
10	36.4399999380112\\
11	37.1399998664856\\
12	37.720000743866\\
13	38.3500009775162\\
14	38.9999985694885\\
15	39.2399996519089\\
16	40.1199996471405\\
17	40.6199991703033\\
18	41.3500010967255\\
19	41.6500002145767\\
20	42.1000003814697\\
21	42.5700008869171\\
22	43.0000007152557\\
23	43.4899985790253\\
24	44.0699994564056\\
25	44.4000005722046\\
26	44.8199987411499\\
27	45.320001244545\\
28	45.4699993133545\\
29	46.0200011730194\\
30	46.4399993419647\\
31	46.560001373291\\
32	47.1100002527237\\
33	47.0699995756149\\
34	47.4500000476837\\
35	48.0599999427795\\
36	48.3999997377396\\
37	48.4899997711182\\
38	48.9300012588501\\
39	48.989999294281\\
40	49.439999461174\\
41	49.210000038147\\
42	49.6100008487701\\
43	49.8600006103516\\
44	49.9700009822845\\
45	50.1299977302551\\
46	50.5400002002716\\
47	50.1699984073639\\
48	50.7600009441376\\
49	51.0699987411499\\
50	51.5500009059906\\
51	51.2199997901917\\
52	51.690000295639\\
53	51.8800020217896\\
54	51.7899990081787\\
55	51.8100023269653\\
56	52.0699977874756\\
57	52.1600008010864\\
58	52.3000001907349\\
59	52.4100005626678\\
60	52.7999997138977\\
61	52.7899980545044\\
62	53.0900001525879\\
63	53.1000018119812\\
64	53.2899975776672\\
65	53.5600006580353\\
66	53.600001335144\\
67	53.8500010967255\\
68	54.1499972343445\\
69	54.5799970626831\\
70	53.8999974727631\\
71	54.6700000762939\\
72	55.0700008869171\\
73	54.9499988555908\\
74	55.1400005817413\\
75	55.2500009536743\\
76	54.9499988555908\\
77	55.57000041008\\
78	55.8300018310547\\
79	55.0999999046326\\
80	55.9400022029877\\
81	56.0100018978119\\
82	55.8600008487701\\
83	56.2799990177155\\
84	56.4999997615814\\
85	56.4400017261505\\
86	56.6399991512299\\
87	56.2799990177155\\
 };
\addlegendentry{FL, $L=5$}
\addplot [color=blue]
 table[row sep=crcr]{%
1	8.87999981641769\\
2	16.7199999094009\\
4	21.9699993729591\\
5	23.9800006151199\\
7	27.3200005292892\\
9	30.1800012588501\\
11	31.810000538826\\
13	33.160001039505\\
15	34.0499997138977\\
19	35.4299992322922\\
23	36.8900001049042\\
27	38.4200006723404\\
31	39.5000010728836\\
36	40.8499985933304\\
39	41.3199990987778\\
42	41.9699996709824\\
45	42.6999986171722\\
48	43.2399988174438\\
50	43.8899993896484\\
52	43.9900010824203\\
56	44.8199987411499\\
60	45.5399990081787\\
64	46.0500001907349\\
68	46.6600000858307\\
72	47.2600013017654\\
75	47.5300014019012\\
77	48.0300009250641\\
79	48.5199987888336\\
81	48.6400008201599\\
83	48.870000243187\\
85	49.1699993610382\\
87	49.3099987506866\\
};
\addlegendentry{FL, $L=2$}
\addplot [color=black, mark=triangle*, loosely dotted, mark options={scale=1.3} ]
table[row sep=crcr]{%
87	56.2799990177155\\
};

\addplot [color=black, mark=diamond, mark options={scale=1.1} ]
table[row sep=crcr]{%
75	55.2500009536743\\
};
\addplot [color=black, mark=square*,mark options={scale=0.8},loosely dotted]
table[row sep=crcr]{%
50	51.5500009059906\\
};

\addplot [color=black, mark=square, mark options={scale=0.6} ]
table[row sep=crcr]{%
1	10.4500003159046\\
};

\addplot [color=black, mark=triangle*, loosely dotted, mark options={scale=1.3} ]
table[row sep=crcr]{%
87	49.3099987506866\\
};
\addlegendentry{$\theta_1=1$, $\theta_2 = 0.01$}
\addplot [color=black, mark=diamond, mark options={scale=1.1} ]
table[row sep=crcr]{%
75	47.5300014019012\\
};
\addlegendentry{$\theta_1=1$, $\theta_2 = 1$}
\addplot [color=black, mark=square*,mark options={scale=0.8},loosely dotted ]
table[row sep=crcr]{%
50	43.8899993896484\\
};
\addlegendentry{$\theta_1=0$, $\theta_2 =1$}
\addplot [color=black, mark=square,mark options={scale=0.6} ]
table[row sep=crcr]{%
1	8.87999981641769\\
};
\addlegendentry{$\theta_1=1$, $\theta_2 = 0$}

\end{axis}
\end{tikzpicture}%
}
\end{minipage}

\caption{ Performance of our approach for different~$\theta_1$ and~$\theta_2$,~$K =20$. 
}
\label{fig: theta_accK20}
\end{figure}
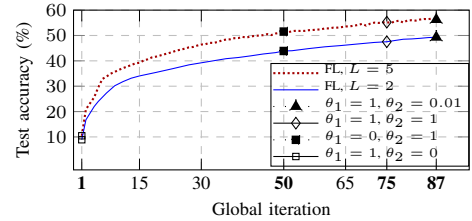


TABLE~\ref{tab: comparison} compares the achieved test accuracy and the corresponding values of stopping iterations~$T=\floor{T_{\max}}$ when applying our power allocation approach, and with two benchmarks: power allocations max-sum rate~\cite{cellfreebook} and the max-min energy-efficiency of generalized Dinkelbach method~\cite{EE_dinkel} for $K =20, 40$,~$L =2, 5$ $\theta_1 = 1$, $\theta_2 = 1$. We observe that for~$K=20$ with equal budget~$\mathcal{L} = 0.2$\,ks, $\mathcal{E}=0.2$\,kJ, our approach outperforms the Dinkelbach approach by up to~$10.5$\% (for $L = 5$) in achievable test accuracy and max-sum rate by up to $20$\%~(for $L = 5$). The similar behavior applies for~$K=40$ with equal budget~$\mathcal{L} = 1$\,ks, $\mathcal{E}=0.5$\,kJ when our approach outperforms the Dinkelbach by up to~$21$\% (for $L = 5$), and the max-sum rate by up to~$27$\% (for $L = 5$) in achievable test accuracy.

Fig.~\ref{fig: theta_accK20} shows the performance of our approach for different values of~$\theta_1$ and~$\theta_2$, which the marks show the achieved test accuracy (with corresponding~$T$) considering~$K=20$, $\mathcal{L} = 0.2$\,ks, $\mathcal{E}=0.2$\,kJ. We observe that solutions where both $\theta_1$ and $\theta_2$ are non-zero results in the highest test accuracy. This outcome emphasizes the significance of our approach, which simultaneously optimizes and strikes a balance between energy and latency, preventing excessive resource utilization in scenarios where power and latency are constrained. 

Fig.~\ref{fig: Accs40N} illustrates the performance of our approach while considering the total AP antennas as~$M\times N=64, 100$ with different values of~$M$ and~$N$. Fig.~\ref{subfig: antenna_cost} shows the total implementation cost, $\bar{C}$, of~$M$ APs with~$N$ antenna per each AP according to the cost model in~\cite{multi_antenna}, as~$\bar{C}~=~M(1~+~c~N)$, where~$c$ is the cost
of installing one antenna at one AP. Fig.~\ref{subfig: K40_FL_M_N} shows the test accuracy of FL over CFmMIMO for~$K=40$,~$\mathcal{L} = 1$\,ks, $\mathcal{E}=0.5$\,kJ,~$\theta_1=1$,~$\theta_2=1$ when applying our power allocation approach and obtain the corresponding values of~$T$ for each scenario of~$M$, and $N$. Although increasing the value $M \times N$ results in a higher~$T_{\max}$ by increasing the data rate, the additional implementation cost is a disadvantage of increasing~$M\times N$ in FL over CFmMIMO.

 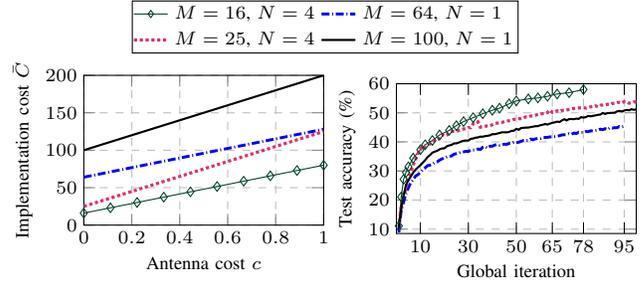
\begin{figure}[t]
 \vspace{5mm}
\centering
\begin{minipage}{0.45\columnwidth}
\vspace{-0.045\textheight}
{\scriptsize\definecolor{amber}{rgb}{1.0, 0.49, 0.0}
\definecolor{taupe}{rgb}{0.28, 0.24, 0.2}
\definecolor{tealgreen}{rgb}{0.0, 0.51, 0.5}
\definecolor{britishracinggreen}{rgb}{0.0, 0.26, 0.15}
\definecolor{cerise}{rgb}{0.87, 0.19, 0.39}
\definecolor{mycolor1}{rgb}{0.00000,0.44700,0.74100}%
\definecolor{mycolor2}{rgb}{0.85000,0.32500,0.09800}%
\definecolor{mycolor3}{rgb}{0.92900,0.69400,0.12500}%
\definecolor{mycolor4}{rgb}{0.49400,0.18400,0.55600}%
\begin{tikzpicture}

\begin{axis}[%
width=0.8\columnwidth,
height=0.5\columnwidth,
at={(0,0)},
scale only axis,
xmin=0,
xmax=1,
ymin=0,
ymax=200,
grid style={dashed},
ymajorgrids,
grid=both,
ylabel near ticks,
ylabel={Implementation cost~$\bar{C}$ },
xlabel={Antenna cost~$c$},
axis background style={fill=white},
legend style={at={(0.2,1.5)},legend columns=2, font = \tiny,  anchor=north west, legend cell align=left, align=left, inner sep=0.01pt, font = \scriptsize,fill=none, text depth=0.25ex, legend image post style={scale=0.75} }
]
\addplot [color=britishracinggreen, mark = diamond, mark options={scale=0.9}]
  table[row sep=crcr]{%
0	16\\
0.111111111111111	23.1111111111111\\
0.222222222222222	30.2222222222222\\
0.333333333333333	37.3333333333333\\
0.444444444444444	44.4444444444444\\
0.555555555555556	51.5555555555556\\
0.666666666666667	58.6666666666667\\
0.777777777777778	65.7777777777778\\
0.888888888888889	72.8888888888889\\
1	80\\
};
\addlegendentry{$M=16$, $N=4$}

\addplot [color=blue,densely dashdotted, line width = 1.1]
  table[row sep=crcr]{%
0	64\\
0.111111111111111	71.1111111111111\\
0.222222222222222	78.2222222222222\\
0.333333333333333	85.3333333333333\\
0.444444444444444	92.4444444444444\\
0.555555555555556	99.5555555555556\\
0.666666666666667	106.666666666667\\
0.777777777777778	113.777777777778\\
0.888888888888889	120.888888888889\\
1	128\\
};
\addlegendentry{$M=64$, $N=1$}

\addplot [color= cerise, densely dotted, line width = 1.2]
  table[row sep=crcr]{%
0	25\\
0.111111111111111	36.1111111111111\\
0.222222222222222	47.2222222222222\\
0.333333333333333	58.3333333333333\\
0.444444444444444	69.4444444444444\\
0.555555555555556	80.5555555555556\\
0.666666666666667	91.6666666666667\\
0.777777777777778	102.777777777778\\
0.888888888888889	113.888888888889\\
1	125\\
};
\addlegendentry{$M=25$, $N=4$}

\addplot [color= black, line width = 0.8]
  table[row sep=crcr]{%
0	100\\
0.111111111111111	111.111111111111\\
0.222222222222222	122.222222222222\\
0.333333333333333	133.333333333333\\
0.444444444444444	144.444444444444\\
0.555555555555556	155.555555555556\\
0.666666666666667	166.666666666667\\
0.777777777777778	177.777777777778\\
0.888888888888889	188.888888888889\\
1	200\\
};
\addlegendentry{$M=100$, $N=1$}

\end{axis}

\end{tikzpicture}%
}
\subcaption{Antenna implementation cost.}
\label{subfig: antenna_cost}
\end{minipage}
\hspace{1mm}
\begin{minipage}{0.45\columnwidth}
{\scriptsize\begin{tikzpicture}
\definecolor{amber}{rgb}{1.0, 0.49, 0.0}
\definecolor{taupe}{rgb}{0.28, 0.24, 0.2}
\definecolor{tealgreen}{rgb}{0.0, 0.51, 0.5}
\definecolor{britishracinggreen}{rgb}{0.0, 0.26, 0.15}
\definecolor{cerise}{rgb}{0.87, 0.19, 0.39}
\begin{axis}[%
width=0.8\columnwidth,
height=0.5\columnwidth,
at={(0,0)},
scale only axis,
xmin=0, 
xmax=100,
xlabel={Global iteration},
xtick={10, 30, 50, 65, 78, 95},
xtick style={color=black},
ymin=8.5,
ymax=60,
ytick={10, 20, 30, 40, 50, 60},
grid style={dashed},
ymajorgrids,
grid=both,
ylabel near ticks,
ylabel={Test accuracy (\%)},
axis background/.style={fill=white},
legend style={at={(1,0.02)}, anchor=south east, legend cell align=left,inner sep=0.01pt, font = \tiny,fill=none, text height=0.4ex, text depth=0.1ex, legend image post style={scale=0.75} }
]

\addplot [color=britishracinggreen, mark = diamond, mark options={scale=0.9}] 
  table[row sep=crcr]{%
1	11.1199997365475\\
2	21.0700005292892\\
3	27.1299988031387\\
4	29.4900000095367\\
5	31.4300000667572\\
7	34.4300001859665\\
10	37.3199999332428\\
12	39.0199989080429\\
15	40.4900014400482\\
18	42.3900008201599\\
22	43.9599990844727\\
25	45.5900013446808\\
28	46.9799995422363\\
32	48.3399987220764\\
36	49.6800005435944\\
40	50.8899986743927\\
43	51.6099989414215\\
47	53.10999751091\\
50	54.1100025177002\\
55	54.7900021076202\\
59	55.1199972629547\\
63	55.6800007820129\\
67	56.4800024032593\\
72	56.9199979305267\\
78	58\\
};

\addplot [color=blue,densely dashdotted, line width = 1.1]
  table[row sep=crcr]{%
1	8.9199997484684\\
2	16.5700003504753\\
3	19.3599998950958\\
4	22.3100006580353\\
5	24.2500007152557\\
6	26.120001077652\\
7	26.9199997186661\\
8	28.2900005578995\\
9	28.7600010633469\\
10	29.5700013637543\\
11	30.3900003433228\\
12	31.0900002717972\\
13	31.8399995565414\\
14	32.0199996232986\\
15	32.3199987411499\\
16	32.9100012779236\\
17	33.1400007009506\\
18	33.7300002574921\\
19	34.1600000858307\\
20	34.4999998807907\\
21	34.799998998642\\
22	35.0800007581711\\
23	35.1599991321564\\
24	35.7600003480911\\
25	35.9600007534027\\
26	36.1299991607666\\
27	36.5299999713898\\
28	36.5599989891052\\
29	36.7700010538101\\
30	36.7300003767014\\
31	37.1800005435944\\
32	37.2099995613098\\
33	37.2000008821487\\
34	37.5400006771088\\
35	37.3800009489059\\
36	38.170000910759\\
37	38.3899986743927\\
38	37.8899991512299\\
39	38.4999990463257\\
40	38.5199993848801\\
41	38.8999998569489\\
42	39.1600012779236\\
43	38.8999998569489\\
44	39.0799999237061\\
45	39.1999989748001\\
46	39.1699999570847\\
47	39.4699990749359\\
48	39.6100014448166\\
49	39.8999989032745\\
50	39.6699994802475\\
51	39.8999989032745\\
52	40.200001001358\\
53	40.4799997806549\\
54	40.4799997806549\\
55	40.5200004577637\\
56	40.9799993038177\\
57	40.7400012016296\\
58	41.0600006580353\\
59	41.2000000476837\\
60	41.1399990320206\\
61	41.5199995040894\\
62	41.5199995040894\\
63	41.5600001811981\\
64	41.7199999094009\\
65	41.7600005865097\\
66	41.839998960495\\
67	41.9600009918213\\
68	42.1400010585785\\
69	42.3900008201599\\
70	42.4600005149841\\
71	42.5399988889694\\
72	42.8200006484985\\
73	42.6099985837936\\
74	43.0599987506866\\
75	42.8799986839294\\
76	43.3400005102158\\
77	43.2700008153915\\
78	43.2500004768372\\
79	43.5799986124039\\
80	43.5099989175797\\
81	43.8199996948242\\
82	43.7099993228912\\
83	43.8899993896484\\
84	43.9999997615814\\
85	44.3899989128113\\
86	44.200000166893\\
87	44.5499986410141\\
88	44.2099988460541\\
89	44.8100000619888\\
90	44.7499990463257\\
91	44.8900014162064\\
92	44.8900014162064\\
93	45.3000009059906\\
94	45.1400011777878\\
95	45.3799992799759\\
};

\addplot [color= cerise, densely dotted, line width = 1.2]
  table[row sep=crcr]{%
1	10.6899999082088\\
2	17.9600004553795\\
3	21.3299993872643\\
4	23.8100001811981\\
5	27.9300003051758\\
6	29.2899994850159\\
7	32.1100004911423\\
8	34.0199995040894\\
9	36.0399988889694\\
10	37.7700008153915\\
11	37.2200009822845\\
12	38.5600007772446\\
13	39.2400003671646\\
14	39.3300004005432\\
15	40.2399994134903\\
16	40.5799992084503\\
17	40.6400002241135\\
18	41.1800004243851\\
19	42.5299988985062\\
20	42.05099985599518\\
21	42.37600013017654\\
22	42.9900007247925\\
23	43.5300009250641\\
24	43.6400012969971\\
25	43.5800002813339\\
26	43.9200000762939\\
27	44.1999988555908\\
28	44.4499986171722\\
29	44.5500003099442\\
30	44.9700014591217\\
31	44.8900001049042\\
32	46.2599989175797\\
33	46.3400002717972\\
34	46.9099994897842\\
35	45.0499988794327\\
36	45.2499992847443\\
37	45.400000333786\\
38	45.4700000286102\\
39	45.4499996900558\\
40	45.9400005340576\\
41	46.3299996852875\\
42	46.4499987363815\\
43	46.5300000905991\\
44	46.6699994802475\\
45	47.0099992752075\\
46	47.1699990034103\\
47	47.2799993753433\\
48	47.4500007629395\\
49	47.7900005578995\\
50	47.7000005245209\\
51	47.9700006246567\\
52	47.9999996423721\\
53	48.2099987268448\\
54	48.609999537468\\
55	48.5899991989136\\
56	48.8999999761581\\
57	48.949999332428\\
58	49.1900004148483\\
59	49.4199998378754\\
60	49.5599992275238\\
61	49.6999986171722\\
62	49.7899986505508\\
63	50.079999089241\\
64	50.9800003767014\\
65	50.1400001049042\\
66	50.2099997997284\\
67	50.7099993228912\\
68	50.419998884201\\
69	50.8400000333786\\
70	50.75\\
71	50.979999423027\\
72	51.1500008106232\\
73	51.3599998950958\\
74	51.4999992847443\\
75	51.5699989795685\\
76	51.6799993515015\\
77	51.6799993515015\\
78	51.7800010442734\\
79	51.5899993181229\\
80	51.8300004005432\\
81	52.0800001621246\\
82	52.1700001955032\\
83	52.4100012779236\\
84	52.0500011444092\\
85	52.6000000238419\\
86	52.6699997186661\\
87	52.7899987697601\\
88	52.9300011396408\\
89	52.7899987697601\\
90	53.2300002574921\\
91	53.2199985980988\\
92	53.5299993753433\\
93	53.5799987316132\\
94	53.3600009679794\\
95	53.849998831749\\
96	53.8100011348724\\
97	53.2999999523163\\
98	53.4399993419647\\
99	53.7599987983704\\
100	53.7099994421005\\
};


\addplot [color= black, line width = 0.8]
  table[row sep=crcr]{%
1	10.5499997735023\\
2	17.0999998450279\\
3	21.5699999332428\\
4	25.2500000596046\\
5	26.8400001525879\\
6	28.1000006198883\\
7	29.4299992918968\\
8	30.4900008440018\\
9	31.2600004673004\\
10	31.9899994134903\\
11	32.8800010681152\\
12	33.7499994039536\\
13	34.7699987888336\\
14	35.4800003767014\\
15	36.2900006771088\\
16	36.6599994897842\\
17	36.92999958992\\
18	37.4600011110306\\
19	37.7300012111664\\
20	37.7899992465973\\
21	38.4800004959106\\
22	38.7799996137619\\
23	38.9800000190735\\
24	39.4900012016296\\
25	39.4999998807907\\
26	39.7699999809265\\
27	40.139998793602\\
28	40.4000002145767\\
29	40.4699999094009\\
30	40.4600012302399\\
31	41.0599994659424\\
32	41.1600011587143\\
33	41.4399999380112\\
34	41.5500003099442\\
35	41.8099987506866\\
36	41.7799997329712\\
37	41.9899988174438\\
38	42.0900005102158\\
39	42.2599989175797\\
40	42.6700013875961\\
41	42.6500010490417\\
42	42.8500014543533\\
43	43.0699992179871\\
44	43.2999986410141\\
45	43.2999986410141\\
46	43.289999961853\\
47	43.4400010108948\\
48	43.7800008058548\\
49	43.850000500679\\
50	44.4800007343292\\
51	44.1699999570847\\
52	44.7700011730194\\
53	44.8400008678436\\
54	44.9100005626678\\
55	44.9300009012222\\
56	45.1100009679794\\
57	45.2399986982346\\
58	45.4000014066696\\
59	45.4199987649918\\
60	45.7299995422363\\
61	45.8899992704391\\
62	46.159999370575\\
63	45.9700006246567\\
64	46.2800014019012\\
65	46.5499985218048\\
66	46.7199999094009\\
67	46.9699996709824\\
68	46.8300002813339\\
69	47.3599988222122\\
70	47.8000003099442\\
71	47.399999499321\\
72	47.8400009870529\\
73	47.8700000047684\\
74	47.8000003099442\\
75	48.0000007152557\\
76	47.9100006818771\\
77	48.5099989175797\\
78	48.3499991893768\\
79	48.529999256134\\
80	48.9200013875961\\
81	48.7999993562698\\
82	48.9599990844727\\
83	49.1199988126755\\
84	49.2999988794327\\
85	49.3899989128113\\
86	49.9299991130829\\
87	49.6200013160706\\
88	49.8900014162064\\
89	49.9499994516373\\
90	50.0100004673004\\
91	50.0100004673004\\
92	50.5000013113022\\
93	50.6499993801117\\
94	50.4800009727478\\
95	50.7899987697601\\
96	50.9899991750717\\
97	50.8700001239777\\
98	51.2399989366531\\
99	51.0299998521805\\
100	51.1600005626678\\
};

\end{axis}
\end{tikzpicture}%
}
\subcaption{FL test accuracy.}
\label{subfig: K40_FL_M_N}
\end{minipage}

\caption{Performance analysis of our approach with corresponding antenna implementation cost, $K=40$,~$\mathcal{L} = 1$ks, $\mathcal{E}=0.5$kJ,~$\theta_1=1$,~$\theta_2=1$.
}
\label{fig: Accs40N}
\vspace{-5mm}
\end{figure}



\section{Conclusion and Future work}
In this paper, we proposed a power allocation scheme to jointly minimize the users' uplink energy and latency for FL over a CFmMIMO network. To this end, we proposed to solve an optimization problem that minimizes a weighted sum of the uplink energy and latency. We have shown that our proposed optimization problem {\color{black}has a unique minimum w.r.t. each user's uplink power} and computed the derivatives in a closed-form solution to enable fast implementation of coordinate gradient descent. The numerical results showed that our proposed power allocation approach outperforms the max-min energy efficiency obtained by the generalized Dinkelbach method and max-sum rate power allocations by respectively achieving up to~$21$\% and~$27$\% increase in test accuracy in energy and latency-constrained situations. Moreover, the numerical results showed that we need to consider the implementation cost of adding APs and antennas in CFmMIMO systems.

For future work, we will consider FL users transmitting different numbers of bits deploying quantization methods and propose novel power allocation schemes in FL over CFmMIMO. Moreover, we will focus on reducing the straggler effect in FL training over CFmMIMO.

\appendices

\section{{Proof of Lemma~\ref{lemma: J_cnx}}\label{P:lemma: J_cnx}}\label{A: proof}


We rewrite the expression of~$\nu(\bp; \theta_1, \theta_2, b, d)$ in~\eqref{eq: nu definition} as
\vspace{-0.01\textheight}
\begin{alignat}{3}\label{eq: nu definition1}
{\nu}(\bp;\theta_1,\theta_2, b, d) = {\psi}_j + \sum_{\substack{\substack{{k'}=1,k' \neq j}}}^K {\psi_{k'}} = {\psi}_j + {\psi}_{[K]\setminus j}.
  \vspace{-0.008\textheight}
\end{alignat}
The first part consists of the weighted sum of energy and latency for a particular user~$j$, and the second part is the summation of the weighted sums over all other users, i.e.,~$k'~\neq~j$. We will first show that~$\psi_j$ is a quasi-convex function of~$p_j$. We define~$\psi_j := F(p_j)/L_g(p_j)$, where~$L_{g}(p_j):= \log_2\left(1 + p_j \bar{A}_j/\left(p_j\bar{B}_j + \bar{C}_j\right)\right)$, and $F(p_j):= \theta_1 p^up_j + \theta_2$. It is easy to see that~$L_g(p_j)$ is a concave function of~$p_j$, $F(p_j)$ is a convex (linear) function of $p_j$, and $1/L_g(p_j)$ is a convex function of~$p_j$. We define~the sublevel set $\mathcal{S}_{\epsilon}$ as $\mathcal{S}_{\epsilon}: = \left \{p_j| F(p_j)/L_g(p_j) \le \epsilon \right\}$, for any $\epsilon \ge 0$, and show that~$\mathcal{S}_{\epsilon}$ is a convex set. Considering 
\begin{equation}\label{eq: S}
   \mathcal{S}_{\epsilon} =  \left \{p_j| F(p_j) - \epsilon L_g(p_j)\le 0 \right\},~0 \le p_j \le 1,
     \vspace{-0.006\textheight}
\end{equation}
where~$-\epsilon L_g(p_j)$ is convex, thus~$\mathcal{S}_{\epsilon}$ is a convex set, and~$\psi_j$ is quasiconvex. Moreover,~$\psi_j$ is differentiable and continuous for~$0 < p_j \le 1$ and has a unique minimum at which the first derivative of~$\psi_j$ is zero, as
\vspace{-0.01\textheight}
\begin{equation}
\partial {\psi}_j|_{p_j = p_{j,1}} := \frac{\partial {\psi}_j}{\partial p_j}|_{p_j = p_{j,1}} = 0.
  \vspace{-0.008\textheight}
\end{equation}
Next, we consider~$\nu_j = {\psi}_j + {\psi}_{[K]\setminus j}$, and its derivative
\vspace{-0.008\textheight}
\begin{equation}\label{eq: psi_partial}
    \partial \nu_j = \frac{\partial {\psi}_j}{\partial p_j} + \frac{\partial {\psi}_{[K]\setminus j}}{\partial p_j}:= \partial {\psi}_j+ \partial {\psi}_{[K]\setminus j}.
      \vspace{-0.008\textheight}
\end{equation}
According to the expressions in~\eqref{eq: p_j^* solution}-\eqref{eq: definitions1}, we obtain that~$\partial {\psi}_{[K]\setminus j} > 0$, which denotes that~$ {\psi}_{[K]\setminus j}$ is an increasing function w.r.t.~$p_j$. Furthermore,~$ {\psi}_{[K]\setminus j}$ is differentiable and continuous for~$0< p_j \le 1$. Thus,~$\nu_j$ is differentiable and~$\partial \nu_j$ is continuous, and we define~$p_{j,2}$ as the smallest $0 \le p_j \le 1$ at which $\partial {\nu_j}|_{p_j = p_{j,2}} = 0$, as
\vspace{-0.008\textheight}
\begin{equation}
  \partial \nu_j|_{p_j = p_{j,2}} =   \partial {\psi}_j|_{p_j = p_{j,2}} + \partial {\psi}_{[K]\setminus j}|_{p_j = p_{j,2}} = 0,
\end{equation}
which results in~$\partial {\psi}_j|_{p_j = p_{j,2}} = - \partial {\psi}_{[K]\setminus j}|_{p_j = p_{j,2}} \le 0$. By investigating the value of~$\nu_j|_{p_j \rightarrow 0} \rightarrow +\infty$, and the continuous behavior of~$\partial \nu_j$, we observe that~$\nu_j|_{p_j = p_{j,2}}$ is a local minimum of~$\nu_j$, because~$p_{j,2}$ is the smallest value at which~$\partial \nu_j = 0$. 

Afterward, since~$\psi_j$ is a differentiable quasiconvex function with continuous derivatives and is non-decreasing for~$p_j \le p_{j,1}$, thus $\psi_j|_{p_j = p_{j,2}} \ge \psi_j|_{p_j = p_{j,1}}$, 
results in~$p_{j,2}~\le~p_{j,1}$. Moreover, $\partial \psi_j|_{p_j > p_{j,1}} \ge 0$, and we have denoted that~$\partial {\psi}_{[K]\setminus j} > 0$, which results in~$\partial \nu_j|_{p_j > p_{j,1}} > 0$. As a result,~$\partial \nu_j \neq 0$ for $ p_j < p_{j,2}$ and~$p_j \ge p_{j,1}$. To investigate the behavior of~$\nu_j$ for $p_{j,2}<p_j< p_{j,1}$, we recall the definition in~\eqref{eq: nu definition}~and~\eqref{eq: nu definition1}, the function~${\psi}_{[K]\setminus j}$ is
\vspace{-0.01\textheight}
\begin{equation}
    {\psi}_{[K]\setminus j} = \sum_{\substack{\substack{{k'}=1\\k' \neq j}}}^K \frac{\theta_1 {\color{black}{p^u}} p_{k'} + \theta_2}{\log_2\left(1 + p_{k'} \bar{A}_{k'}/\left(p_j\Tilde{B}_{k'}^{j} + \Tilde{C}_{k'}^{j}\right) \right)},~\forall j\in[K],
     \vspace{-0.005\textheight}
\end{equation}
where~${\psi}_{[K]\setminus j}$ is a concave function w.r.t.~$p_j$. Depending on the value of the second derivative of~${\psi}_{[K]\setminus j}$, defined as~${\partial}^2 {\psi}_{[K]\setminus j}$, there is either no local optimum or an equal number of local minimum and local maximum in~$\nu_j$, for~$p_{j,2}<p_j<p_{j,1}$. In this paper, the value of~${\partial}^2 {\psi}_{[K]\setminus j}$, for~$p_{j,2}<p_j<p_{j,1}$, is negligible compared to the second derivative of~${\partial}^2 \psi_j$ (the analytical argument is long and out of the scope of this paper). Therefore, ${\partial}^2 \nu_j = {\partial}^2 \psi_j + {\partial}^2 {\psi}_{[K]\setminus j} \ge 0$, results in~$p_j^* = p_{j,2}$, which shows we obtain a unique minimum for~$\nu_j$.


    
 
\bibliographystyle{./MetaFiles/IEEEtran}
\bibliography{./MetaFiles/References}

\begin{thebibliography}{10}
\providecommand{\url}[1]{#1}
\csname url@samestyle\endcsname
\providecommand{\newblock}{\relax}
\providecommand{\bibinfo}[2]{#2}
\providecommand{\BIBentrySTDinterwordspacing}{\spaceskip=0pt\relax}
\providecommand{\BIBentryALTinterwordstretchfactor}{4}
\providecommand{\BIBentryALTinterwordspacing}{\spaceskip=\fontdimen2\font plus
\BIBentryALTinterwordstretchfactor\fontdimen3\font minus \fontdimen4\font\relax}
\providecommand{\BIBforeignlanguage}[2]{{%
\expandafter\ifx\csname l@#1\endcsname\relax
\typeout{** WARNING: IEEEtran.bst: No hyphenation pattern has been}%
\typeout{** loaded for the language `#1'. Using the pattern for}%
\typeout{** the default language instead.}%
\else
\language=\csname l@#1\endcsname
\fi
#2}}
\providecommand{\BIBdecl}{\relax}
\BIBdecl

\bibitem{konevcny2016federated}
J.~Kone{\v{c}}n{\`y} \emph{et~al.}, ``Federated {L}earning: {S}trategies for improving communication efficiency,'' \emph{arXiv preprint arXiv:1610.05492}, 2016.

\bibitem{cellfreebook}
O.~T. Demir \emph{et~al.}, ``Foundations of user-centric cell-free {M}assive {MIMO},'' \emph{Foundations and Trends in Signal Processing}, vol.~14, no. 3-4, pp. 162--472, 2021.

\bibitem{9311931}
M.~Chen \emph{et~al.}, ``Wireless communications for collaborative {F}ederated {L}earning,'' \emph{IEEE Communications Magazine}, vol.~58, pp. 48--54, 2020.

\bibitem{9796621}
T.~T. Vu \emph{et~al.}, ``Joint resource allocation to minimize execution time of {F}ederated {L}earning in cell-free {Massive MIMO},'' \emph{IEEE Internet of Things Journal}, vol.~9, no.~21, pp. 21\,736--21\,750, 2022.

\bibitem{9975256}
J.~Zhang \emph{et~al.}, ``{F}ederated {L}earning-based cell-free {Massive MIMO} system for privacy-preserving,'' \emph{IEEE Transactions on Wireless Communications}, vol.~22, no.~7, pp. 4449--4460, 2023.

\bibitem{9500541}
T.~T. Vu \emph{et~al.}, ``Straggler effect mitigation for {F}ederated {L}earning in cell-free {Massive MIMO},'' in \emph{ICC 2021 - IEEE International Conference on Communications}, 2021, pp. 1--6.

\bibitem{9124715}
------, ``Cell-free {Massive MIMO} for wireless {F}ederated {L}earning,'' \emph{IEEE Transactions on Wireless Communications}, vol.~19, no.~10, pp. 6377--6392, 2020.

\bibitem{8781848}
M.~Bashar \emph{et~al.}, ``Energy efficiency of the cell-free {Massive MIMO} uplink with optimal uniform quantization,'' \emph{IEEE Transactions on Green Communications and Networking}, vol.~3, no.~4, pp. 971--987, 2019.

\bibitem{vu2021does}
T.~T. Vu \emph{et~al.}, ``How does cell-free {Massive MIMO} support multiple {F}ederated {L}earning groups?'' in \emph{2021 IEEE 22nd International Workshop on Signal Processing Advances in Wireless Communications (SPAWC)}, 2021, pp. 401--405.

\bibitem{10120750}
T.~Zhao \emph{et~al.}, ``Energy-efficient {F}ederated {L}earning over cell-free {IoT} networks: Modeling and optimization,'' \emph{IEEE Internet of Things Journal}, pp. 1--1, 2023.

\bibitem{EE_dinkel}
A.~Zappone \emph{et~al.}, ``Energy-efficient power control: {A} look at {5G} wireless technologies,'' \emph{IEEE Transactions on Signal Processing}, vol.~64, pp. 1668--1683, 2015.

\bibitem{li2019convergence}
X.~Li \emph{et~al.}, ``On the convergence of {F}ed{A}vg on non-iid data,'' \emph{arXiv preprint arXiv:1907.02189}, 2019.

\bibitem{8889996}
F.~Sattler \emph{et~al.}, ``Robust and communication-efficient {F}ederated {L}earning from non-i.i.d. data,'' \emph{IEEE Transactions on Neural Networks and Learning Systems}, vol.~31, no.~9, pp. 3400--3413, 2020.

\bibitem{7827017}
H.~Q. Ngo \emph{et~al.}, ``Cell-free {Massive MIMO} versus small cells,'' \emph{IEEE Transactions on Wireless Communications}, vol.~16, pp. 1834--1850, 2017.

\bibitem{boydcnvx}
S.~Boyd \emph{et~al.}, \emph{Convex Optimization}.\hskip 1em plus 0.5em minus 0.4em\relax USA: Cambridge University Press, 2004.

\bibitem{multi_antenna}
A.~A.~I. Ibrahim \emph{et~al.}, ``Cell-free massive {MIMO systems} utilizing multi-antenna access points,'' in \emph{2017 51st Asilomar Conference on Signals, Systems, and Computers}, 2017, pp. 1517--1521.

\end{thebibliography}
\end{document}